# Co-evolutionary hybrid intelligence

Kirill Krinkin
*Alexander Popov International Innovation Institute for AI, Cybersecurity and Communications, SPbETU "LETI"*
St. Petersburg, Russia
kirill@krinkin.com

Yulia Shichkina
*Alexander Popov International Innovation Institute for AI, Cybersecurity and Communications, SPbETU "LETI"*
St. Petersburg, Russia
shichkina@etu.ai

Andrey Ignatyev
*MGIMO University Center for Global IT Cooperation (CGITC)*
Moscow, Russia
andrey.ignatyev.g@gmail.com

*Abstract*— Artificial intelligence is one of the drivers of modern technological development. The current approach to the development of intelligent systems is data centric. It has several limitations: it is fundamentally impossible to collect data for modeling complex objects and processes; training neural networks requires huge computational and energy resources; solutions are not explainable. The article discusses an alternative approach to the development of artificial intelligence systems based on the human-machine hybridization and their co-evolution.

*Keywords*— artificial intelligence concepts, intelligent hybrid systems, co-evolution, AI Ethics, human-machine interoperability hybrid intelligence evaluation.

## I. INTRODUCTION

Artificial intelligence (AI) is a very broad concept that developers use in a variety of industries: from computer games to voice assistants such as Alice or Siri. There is no universally accepted international definition. Some experts have an opinion that such situation is fully justified and permissible. The two main basic approaches are AI as a scientific direction and AI as a complex of technologies.

The global trend for artificial intelligence development has been set during Dartmouth seminar in 1956. The main goal was to define characteristics and research directions for artificial intelligent comparable or even overperformed than human intelligence. It should be able to acquire and create new knowledges in highly uncertain dynamic environment (the real-world environment is an example) and apply those knowledges for solving practical problems.

Nowadays, the AI-systems (based on narrow AI [1]) which are being developed, can hardly be considered intelligence. It is rather the next level of automation of human work.

The main characteristics of a strong AI are:

- the ability to independently formalize tasks, change (create) algorithms; the ability to change ontologies;
- comprehensive meaning processing;
- causal relations identification [2];
- human-machine and machine-machine interoperability (the ability to perform joint work, building division of a labor).

Interoperability provides such properties as:

- explainability;
- operating in a shared (and open global) ontology and the ability to transfer the context of a task between subjects of actions;
- ability to transfer of knowledge and experience from one context (ontology) to another on the fly

It is not the purpose of this paper to find the most relevant definition of AI. Further, in the paper, we will use on the following definition of artificial intelligence.

AI is an intelligence implemented on an artificial substrate (machine), where *the Intelligence* is a measurable property of a system in the ability to acquire, accumulate, and use knowledge, based on the previous experience of solving problems (it does not matter whom experience has been used for knowledge extraction)

Today, challenges are related to the complexity of a human-created systems (social, technical, economic, political and others ...) and natural processes (or consequences of human actions in the biosphere, such as pandemics or climate changes). Human intelligence equipped existing intellectual tools is unable to cope with it and realize all the interconnections and inter-dependencies. Significantly more data is required to create models of complex objects based on traditional AI approaches. At the same time, the existing computing capabilities are practically exhausted. Consequently, there is a fundamental impossibility to build an adequate model of complex objects and processes based on data-centered AI. It is impossible to manage those objects without models. It means that it is impossible to consciously shape the future. This contradiction between the incomprehensible complexity of systems and the need to manage them opens a window of opportunity for the expansion of human intelligence by machine (artificial intelligence).

Instead of discussing strong artificial intelligence development it is better to fucus on *intelligence* itself and its properties. After that it is necessary to clarify a set those of cognitive functions which can be transferred to an artificial substrate. The purpose of this article is to show a possible way to create strong intelligent systems based on the hybridization of artificial and human capabilities and their co-directional evolution.

## II. CO-EVOLUTIONARY HYBRID INTELLIGENCE CONCEPT

The call for awareness of the "human side of computing" has been repeated since the mid-1980s. J. Grudin, for example, noted that Computer effectiveness grows with the increasing their understanding of humans [3].

D. Engelbart [4] showed that designing the interface of the personal computer was a technology of people and about people. Engelbart views the relationship between people and programs as a heterogeneous community in which the evolution of all agents involved in the system takes place. Inside complex modern information systems, a joint evolution is taking place, in which people and means are involved. It is already a mutually enriching digital ecosystem. He noted that there is a constant joint process of co-evolutionary changes between technical means and the people who use them. In any situation there are interrelated elements: instruments of activity, activity and people as subjects of activity. Changes in each of the interrelated elements that make up the socio-technical system of activity lead to changes in other elements. As soon as new technical means of activity are created and used, they change the usual conditions of existence. and new forms of joint activity begin to form around them. It causes the people begin to master new roles, develop new habits and master new methods, processes and behaviors.

The following four areas need to be taking into account in order to organize the development, implementation and introduction of strong intelligence:

- global challenges, problems, and tasks that contribute to the emergence of new products in the field of AI on the market;
- products, which are various intelligent systems and their components. Market needs define technologies and drive their development;
- AI technologies that are based on existing concepts of building AI. Technologies can include both various methods and fundamental theories;
- novel concepts as fundamental theories that allow for new breakthrough solutions in the field of AI.

Each of them has been developed independently. In order to reach a consistency of intelligent systems it is necessary to provide synchronization and connection between these areas.

The development of the intelligence in general (including the strong artificial intelligence) cannot be carried out without a system of criteria. One of these criteria is a benchmark allowing to evaluate the level of the intelligence and a superiority to human capabilities. F. Chollet [5] suggested to measure the intelligence rely on ability to generalization and experience transfer. In case of strong intelligence two approaches can be distinguished:

- a synthetic test, an analogue of the Turing test, which confirms the given characteristics;
- the solution of the global, objectively existing problem that the collective intelligence of a person is not able to solve.

The human cannot create something exceeding his cognitive abilities. Thereby, the analog of the Turing test for the strong intelligence can be created.

Consequently, solving a problem that human intelligence is unable to solve, may be a good benchmark. To accept the fact the problem is solved the human intelligence must be developing synchronously with artificial one.

In other words, a hypothetical model of the strong artificial intelligence can only be hybrid. The reasons are following.

- *Sharing of knowledge.* Until the problem is solved the knowledge about the problem and intellectual abilities that allow interpreting this knowledge are distributed between human and machine
- *Subjecness of human.* To state the fact of the solution of the original problem the human should participate in the problem definition, have the intention to solve it, have a vision of the evaluation procedure.

- *Meaninglessness of the solution without human subjectness.* The lost of ability to define the problem, intention to solve, or ability to check the result causes the inability to check the fact of the solution.

The development of intelligence in general, as joint development of natural and artificial intelligence, should occur in such a way that the machine part of the intellect remains comprehensible to the human Therefore both parts would inevitably increase their complexity (that is, continue developing).

Consequently, the process of the development of intelligence in general should correspond to the property of increasing cognitive interoperability. The level of the cognitive interoperability is arranged in the same way as the method of measuring the strength of intelligence according to Chollet [5]. The level increase is not associated with the amount or speed of data transfer between intelligence on different substrate and the possibility of cooperation at higher levels of interaction.

According to Z. Akata et. al [5] the hybrid intelligence (HI) is a combination of human and machine intelligence, augmenting human intellect and capabilities instead of replacing them and achieving goals that were unreachable by either humans or machines. in order to point out to the approach to the development of strong intelligence, this definition has to be extended. It is important to focus on two crucial characteristics:

- interoperability;
- human-machine co-evolution.

In this case, interoperability means the ability of two or more systems to exchange knowledge and use it correctly to solve the problem.

There are several classes of interoperability:

- between the developer and the intelligent system being created;
- between the intelligent system and its users;
- between intelligent systems.

Thereby, **Co-evolutionary hybrid intelligence** (CHI) is a symbiosis of artificial and natural intelligence, mutually developing, teaching, and complementing each other in the process of co-evolution. Human-machine intelligence co-evolution is the fundamental basic process of building of strong intelligent systems.

The speed of developing and implementing hybrid intelligence solutions depends on the following factors:

- cognitive functions formalization level (advancement in the understanding of human intelligence);
- performance of the experience/knowledge transfer from a person to a machine and vice versa;
- new products integration simplicity into the existing system of division of labor and value chains in the market.

Recent results in modeling human cognitive functions and the mathematical view on attention and consciousness promise new approaches to human-machine interoperability. In particular, the Attention Schema Theory [6] and Integrated Information Theory [7]. can become the basis for cognitive interoperability.

The differences between the evolutionary approach to the development of intelligence from the traditional one:

- the hybrid human-instrument system is being developed as a system a whole, but not as the AI as a tool;
- in a hybrid system, there is no explicit dependence on the cognitive function implementation (it can be a human or a method, for example, based on machine learning);
- a hybrid system can train new users, as it accumulates the best user (agent) practices for its application.

The next two sections provide preliminary considerations on the ethics of building hybrid systems and an applied example.

### III. Ethical concerns

Assuming that AI is currently the most data-centric technology capable of integrating elements of other cross-cutting digital technologies, ethical issues in relation to HI or intelligent hybrid systems should probably be considered in a broader sense and scope. When talking about Ethics in relation to HI, we are essentially talking about techno-ethics and philosophy in the field of AI. From the balanced and most universal definition of Ethics in AI as being "a set of values, principles, and techniques that employ widely accepted standards of right and wrong to guide moral conduct in the development and use of AI technologies" [9], it can be assumed that ethical issues in HI are important primarily in relation to a field such as:

- moral standards for deepening human-machine interaction;
- the positive and negative aspects of HI development, its impact/influence on the individual, society, planet and humanity as a whole (issues of HI and civilization trajectory).

This very broad problematic includes issues of risk assessment, i.e., all forms of harm and damage to various kinds of living and non-living objects, above all to humans, including their autonomy, cognitive functions, issues of human freedoms and rights, discrimination, social communication, etc. In this broad direction, in addition to the concerns that characterize the current discourse in AI Ethics, the following aspects are expected to be of especial relevance.

A significant issue for ethical research would be the potential for the human-machine interaction process to get out of control and become unsupervised, where the role and importance of AIS in the bilateral overall process may become asymmetrically reinforced and acquire the properties of a «defining dominant». In other words, by outsourcing to a machine functions that in the mathematical and algorithmic dimension humans are incapable of performing, there is a risk of gradual loss of control not only over a particular system decision, but also of control over strategic goals, the system of checkpoints ("taboos") and the definition of basic movement coordinates that have always been in human hands when interacting with machinery. In simple terms, the problem can be formulated as a model «master and servant» or «captain and helmsman». Even the conditional "equality" model in final decision-making can probably carry with it certain risks of loss of control for the human.

When thinking about the possibilities of HI, it is possible to simulate a situation where a person, using his habitual mode of mental reflection, "plays" in his mind the most egoistic and socially unacceptable scenarios of his behavior. Such "experimental" versions of behaviour when possibly tested and "run-in" in HI scenarios can increase the risks of "real-life use" of socially dangerous schemes and decisions. Unlike, for example, nuclear power management or cloning, in a coevolutionary dialogue with an AI, it will be more difficult for humans to set up barriers and inhibitions, to provide a "safety valve" and a reliable mode of blocking morally inappropriate machine decisions and actions.

Thus, on the one hand, in order to empower algorithms, human will have to expand the autonomy of AIS, on the other hand, to remain the unconditional supervisor and manager of the most complex interaction processes. For the human being himself such a blockage is moral norms, his conscience, ethical and moral orientations (they are formed by family, society, religion, tradition, etc.). With sufficiently tangible breakthrough in autonomy of machines, it seems to be rather difficult to establish and build into the algorithms such "constraints".

As a separate and very important issue in relation to HI it is worth highlighting the risk of cognitive degradation in humans. It can be assumed that by increasingly freeing itself from some of the mental effort and operations that can be delegated to a machine, the human brain can be subjected to changes, both positive and negative, that are already confirmed at this stage of technological development.

Another aspect, which can be designated as significant and specific to HI, is the possibility of HI's "reverse influence" on the "picture of the world" inside a person, on the further development of ethical and moral concepts of a person. These concepts in the wider historical dimension are definitely flexible and mobile, including through the adoption of new technological tools.

In general, the emerging ethical dilemmas in the field of HI cannot be solved at the level of universal approaches and principles; it can only be achieved through the substantive and systematic study of specific systems in specific application environments, which will require both social and philosophical understanding and the development of new highly practical tools for measuring, testing and conducting complex experimental work of an interdisciplinary nature.

Within ethical issues there are still many unresolved problems related to human self-understanding. In this sense, the study of HI could contribute to some extent to solving this primary problem.

### IV. Use Case

Consider an example from the applied medicine. The one of the main problems in this domain is the lack of interoperability between medical intelligent systems (MIS) and physicians. The following evolution stages can be defined for such systems.

At the initial stage, the user can receive only primary data about the object (patient). Such data can be images, data from sensor devices, radiography and so on. The user can label this data and receive a dataset for training the system. Also, the system can use machine learning and classify data. Features in the data that a human cannot classify can be highlighted. At this stage, the system is trained in order to extract hidden features.

At the next stage, these found features can be supplemented with descriptions. Domain experts generate additional meta-data. At the same stage, a group of specialists will mark up the dataset, considering the results of the analysis of new features.

At next stage, the AI system learns classifying objects or making decisions or using this new input data for detecting new features. The process is repeated recursively. Specialists use this AI-decisions for building new concepts in the domain and prepare new input data for AI.

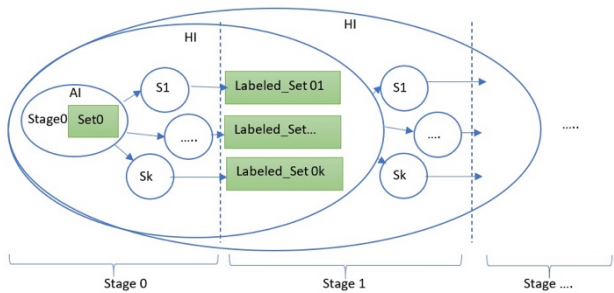

Fig 1. – Co-evolution example, where HI - hybrid intelligence system, Si – domain specialist, Set0 – initial dataset, Labeled_Set01 – labeled dataset.

As an example, we can consider a system for assessing the level of stress based on the analysis of involuntary movements of the movable links of the human limbs. The device for non-invasive measurement of such movements was developed at Saint Petersburg Electrotechnical University. It generates signals that enter the AI system, and it builds the plot. The system detects the characteristic fragments of limbs movements plot of person under certain conditions and a neurological disease.

At the next stage, specialists evaluate these characteristic fragments of the plot and link them with disease stages, treatment periods and physiological characteristics of the human body. The treatment is being adjusted. Based on this data the new diagnosis methods for neurological diseases are being developed. It helps acquiring new knowledge in the neurology domain and developing the human intelligence. Labeled data used for re-training AI-system and support its development.

## CONCLUSION AND FUTURE WORK

A paradigm shift is needed to build intelligence beyond human capabilities. The development of data-centered intelligence is approaching its limits. Instead of a data-centered approach it is required to use intelligence-centered approach. Hybridization of human and machine intellectual capabilities based on cognitive interoperability and co-evolution is a new frontier. To build systems based on hybrid intelligence, the following developments are required:

- methodology for constructing co-evolution systems;
- languages for describing cognitive functions and describing dynamic intelligence: [10];
- frameworks for building hybrid intelligent systems;
- novel technological standards, best practices, and models for hybrid systems;
- ethical specific concepts, approaches and norms to regulate development and use hybrid systems in the human life.

ACKNOWLEDGMENT

This work was supported by the Ministry of Science and Higher Education of the Russian Federation by the Agreement № 075-15-2020-933.